\pdfoutput=1

\documentclass[11pt]{article}

\usepackage{EMNLP2023}

\usepackage{times}
\usepackage{latexsym}

\usepackage[T1]{fontenc}

\usepackage[utf8]{inputenc}

\usepackage{microtype}

\usepackage{inconsolata}

\usepackage{graphicx}
\usepackage{inconsolata}
\usepackage{CJK}
\usepackage{makecell}
\usepackage{multirow}
\usepackage{longtable}
\usepackage{algorithm}
\usepackage{algorithmic}

\usepackage{float}
\usepackage{subcaption}
\usepackage{tabularx}
\usepackage{booktabs}

\title{Proposition from the Perspective of Chinese Language: A Chinese Proposition Classification Evaluation Benchmark}

\author{Conghui Niu\textsuperscript{1}, Mengyang Hu\textsuperscript{1}, Lin Bo\textsuperscript{1}, Xiaoli He\textsuperscript{1}, Dong Yu\textsuperscript{1}, Pengyuan Liu\textsuperscript{1,2}\\
1.Beijing Language and Culture University\\
2.National Language Resources Monitoring and Research Center for Print Media\\
\texttt{niuconghui2001@126.com, yudong@blcu.edu.cn, liupengyuan@pku.edu.cn}}

\begin{document}
\maketitle
\begin{abstract}
Existing propositions often rely on logical constants for classification. Compared with Western languages that lean towards hypotaxis such as English, Chinese often relies on semantic or logical understanding rather than logical connectives in daily expressions, exhibiting the characteristics of parataxis. However, existing research has rarely paid attention to this issue. And accurately classifying these propositions is crucial for natural language understanding and reasoning.
In this paper, we put forward the concepts of explicit and implicit propositions and propose a comprehensive multi-level proposition classification system based on linguistics and logic. Correspondingly, we create a large-scale Chinese proposition dataset PEACE\footnote{The dataset will be publicly released later.} from multiple domains, covering all categories related to propositions.
To evaluate the Chinese proposition classification ability of existing models and explore their limitations, We conduct evaluations on PEACE using several different methods including the Rule-based method, SVM, BERT, RoBERTA, and ChatGPT. Results show the importance of properly modeling the semantic features of propositions. BERT has relatively good proposition classification capability, but lacks cross-domain transferability. ChatGPT performs poorly, but its classification ability can be improved by providing more proposition information. Many issues are still far from being resolved and require further study.
\end{abstract}

\section{Introduction}

The foundation of human research on logic lies in propositions.
As the smallest unit of logical reasoning, propositions are deﬁned as the meaning of declarative sentences in linguistics and logic \citep{sep-propositions}.
According to the characteristics of propositions, it corresponds to different inference rules to complete the research on inter-sentence logic and reasoning.
In logic, logical constants refer to the invariant part of a logical form. Two important types of logical constants are logical connectives and quantifiers. For example, "\textit{All... are...}" in the simple proposition "\textit{All metals are conductive}" and "\textit{If... then...}" in the compound proposition "\textit{If it rains, then the ground is wet}".
These logical constants are usually regarded as the basis for classifying propositions.

In terms of NLP, proposition research is relevant for many downstream tasks. 
One such application is the research of modal proposition, which can aid in the automatic detection of intention, uncertainty, behavior and so on \citep{zerva2017using,vincze2008bioscope,prieto2020data}.
Besides, non-modal proposition is related to logical symbol recognition and logical relationship extraction in sentences
\citep{wang-etal-2022-logic,jiao-etal-2022-merit,huang-etal-2021-dagn}.
Syllogism is the most common form of deductive reasoning, among which categorical syllogism is the most common type. Classifying categorical proposition helps to test the inferential validity of syllogism \citep{fan2016fomal}.
These all emphasize the importance of propositional research in NLP.

\begin{table}[t]
    \centering
    \resizebox{\linewidth}{!}{
    \begin{tabular}{cc}
         \hline
         \textbf{Sentence} & \textbf{Type} \\
         \hline
         \makecell[l]{\begin{CJK}{UTF8}{gbsn}(所有)人(都)是动物。\end{CJK} \\ \textit{\textbf{All} humans \textbf{are} animals.}} & \makecell{Categorical\\Proposition} \\
         \hline
         \makecell[l]{\begin{CJK}{UTF8}{gbsn}(如果)他不来，(那么)我来。\end{CJK} \\ \textit{\textbf{If} he doesn't come, \textbf{then} I'll come.}} & \makecell{Hypothetical\\Proposition} \\
         \hline
         \makecell[l]{\begin{CJK}{UTF8}{gbsn}他喜欢读书，(以及)我喜欢画画。\end{CJK} \\ \textit{He likes reading, \textbf{and} I like drawing.}} & \makecell{Conjunctive\\Proposition} \\
         \hline
    \end{tabular}
    }
    \caption{Examples of Chinese expressions that omit logical constants. Contents in parentheses are omitted logical constants}
    \label{tab:omited propositions}
\end{table}

Western languages such as English often use and rely on various connectives that can express logical relationships in their expressions, showing the external features of hypotaxis.
However, Chinese, as a language that leans towards parataxis, has the characteristics of simplicity (i.e. linguistic components can be omitted as long as they mean idea can get across) and flexibility (i.e. there is no fixed correspondence between parts of speech and syntactic components; the same word can function as different syntactic components)\citep{zhang2001philosophical,tse2010parataxis,yu1993chinese,shi2000flexibility}.
It often relies on semantic or logical understanding rather than connectives in natural language expression\citep{cao2002logical,zhang2001philosophical}.
As shown in Table~\ref{tab:omited propositions}, although many expressions in Chinese omit logical constants, their logical content is very clear. Previous research on propositions didn't pay attention to this issue and lacked corresponding corpus resources.

In this paper, we propose the concepts of explicit and implicit proposition, break the limitations of logical constants and expand the research scope of proposition to actual Chinese natural language. We refer to propositions containing complete logical keywords (e.g., "if... then...", "all... are...", "must") as explicit propositions. On the contrary, propositions that don't fully appear the above features at the linguistic level but are implicit in semantic logic are called implicit propositions.
Additionally, we introduce a comprehensive multi-level proposition classification system based on linguistics and logic, covering all categories related to propositions.
We accordingly construct a large-scale Chinese proposition dataset PEACE containing over 45k sentences from different domains.
Large language models represented by ChatGPT have made remarkable progress in natural language processing and artificial intelligence\citep{scao2022bloom,wei2022emergent}. This inspires us to evaluate the Chinese proposition classification ability of existing models and explore their limitations.
We conduct evaluations on PEACE using several different methods including the Rule-based method, SVM, BERT, RoBERTA, and ChatGPT.
Experiments show that BERT has relatively good proposition classification capability, but lacks cross-domain transferability. ChatGPT excels at capturing explicit rather than implicit propositional features. Its classification ability can be improved by providing more proposition information.

Our contributions are summarized as follows:
\begin{itemize}
    \item Based on the characteristics of Chinese, we put forward the concepts of explicit and implicit propositions, which is more suitable for Chinese NLP scenarios.
    \item We propose a comprehensive multi-level proposition classification system and create a large-scale Chinese proposition dataset from multiple domains, covering all categories related to propositions.
    \item We evaluate the performance of multiple models in in-domain and cross-domain proposition classification on PEACE. Results show the importance of properly modeling the semantic features of propositions. These issues are still far from being resolved and require further study.
\end{itemize}

\section{Related Work}

\citet{liu2021automatic} first introduced the concept of Chinese proposition into NLP.
They use logical constants as the basis for classification to determine whether a natural language sentence is a proposition, limited to the study of explicit proposition.
\citet{hu2021propc} constructed a Chinese dataset ProPC for in-domain and cross-domain non-modal proposition classification. They focus on a part of the proposition, namely the non-modal proposition.
And there is some overlap between different domains.
\citet{pyatkin-etal-2021-possible} proposed modal sense hierarchical classification based on modal event detection and verified the improvement of modal event detection by modal classification.
But the taxonomy they proposed is too fine-grained. For example, for the alethic modality based on objective world states (\textit{e.g. Water heated to boiling point will \textbf{inevitably} become water vapor}) and the cognitive modality based on personal subjective experience and knowledge (\textit{e.g. I think Mr. Wang \textbf{must} be at school}) that both express possibility, they explained the difference between them by "by state of word" and "by state of knowledge".
It increases the cost of manual annotation. And there's no close connection between such fine-grained classification and downstream tasks. In our work, these two are merged into one category in the NLP.

\begin{figure*}[t]
    \centering
    \includegraphics[width = \linewidth]{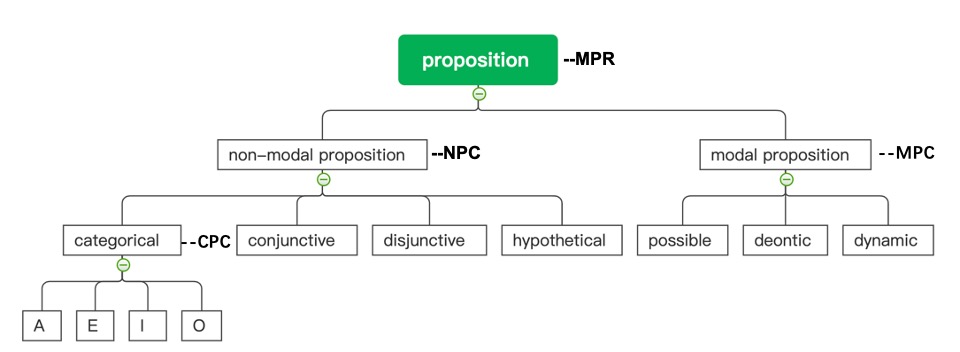}
    \caption{The proposed framework of proposition classification and the corresponding relationship with our tasks.
    }
    \label{fig:framework}
\end{figure*}

\section{Proposition Definition and Benchmark Tasks}
\label{sec:3}

Figure~\ref{fig:framework} presents the taxonomy we proposed for proposition classification in NLP based on linguistics and logic.
Specific examples are shown in Appendix ~\ref{sec:appendix}.

\subsection{Definitions}

\noindent\textbf{Non-modal Proposition}
A deterministic judgment on the actual existence or non-existence of things.
\begin{itemize}
    \item \noindent\textbf{Categorical Proposition} Make a direct and unconditional judgment on whether an object logically contains a certain attribute or belongs to a certain category.
    \item \noindent\textbf{Conjunctive Proposition} A compound proposition, which reflects that several situations or properties of objects exist at the same time, and logically has a conjunctive relationship. 
    \item \noindent\textbf{Disjunctive Proposition} A compound proposition, which reflects the existence of at least one condition or attribute of an object, and logically has a disjunctive relationship.
    \item \noindent\textbf{Hypothetical Proposition} A compound proposition, which contains a previous or tentative explanation, and logically has a conditional relationship.
\end{itemize}

\noindent\textbf{Categorical Proposition}
According to the difference of quantity term and quality term, the property proposition can be divided into the following four types: \footnote{The letters "A" and "I" came from the Latin affirmo (I affirm), while "E" and "O" from the Latin nego (I deny).}
\begin{itemize}
    \item \textbf{Universal Affirmative Proposition (A):} A proposition that semantically concludes that all objects of a class of things have certain properties.\footnote{Simple proposition refers to only one object, so it can generally be regarded as a universal proposition.} The explicit proposition has the form: "\textit{All} S \textit{are} P".
    \item \textbf{Universal Negative Proposition (E):} A proposition that semantically concludes that all objects of a class of things have no certain properties. The explicit proposition has the form: "\textit{No} S \textit{is} P".
    \item \textbf{Particular Affirmative Proposition (I):} A proposition that semantically concludes that some objects of a class of things have certain properties. The explicit proposition has the form: "\textit{Some} S \textit{are} P".
    \item \textbf{Particular Negative Proposition (O):} A proposition that semantically concludes that some objects of a class of things have no certain properties. The explicit proposition has the form: "\textit{Some} S \textit{are not} P".
\end{itemize}

\noindent\textbf{Modal Proposition}
Modality refers to the speaker's modification of state of affairs, which is used to express the concepts of possibility, inevitability, promise, obligation and ability.
\begin{itemize}
    \item \noindent\textbf{Possible Modal} The speaker makes a decision about the likelihood that the central meaning expressed by the proposition will occur.
    \item \noindent\textbf{Deontic Modal} The speaker allows or gives instructions that make actions, states, and events possible or to be performed.
    \item \noindent\textbf{Dynamic Modal} Focus on the subject, participant's ability or willingness, not the speaker's point of view or attitude.
\end{itemize}

\subsection{Task Formulation}

\noindent \textbf{Modal Proposition Recognition (MPR)}
Concretely, proposition is first divided into non-modal proposition and modal proposition based on whether a deterministic judgment is made on the object \citep{sep-propositions,fan2016fomal}.
Modal proposition hasn't been paid enough attention in previous studies of NLP, which leads to many modal propositions being incorrectly labeled as non-modal propositions.
Therefore, it is necessary to identify modal proposition to distinguish it from non-modal proposition.
This task is to predict whether a given natural language sentence is a proposition and whether it is a modal proposition. We define it as a three-category task (\textit{not-proposition/non-modal/modal}).
\\
\\
\noindent \textbf{Non-modal Propositions Classification (NPC)}
Non-modal proposition is further divided into simple proposition and compound proposition.
Simple proposition is also known as categorical proposition.
Compound proposition consists of conjunctive proposition, hypothetical proposition and disjunctive proposition \citep{fan2016fomal}. 
This task is to predict the category of a given non-modal proposition. We define it as a four-category task (\textit{categorical/conjunctive/hypothetical/disjunctive}).
\\
\\
\noindent \textbf{Categorical Proposition Classification (CPC)}
Categorical proposition consists of universal affirmative proposition (A), universal negative proposition (E), particular affirmative proposition (I), and particular negative proposition (O).
This task is to predict the category of a given categorical proposition. We define it as a four-category task (\textit{A/E/I/O}).
\\
\\
\noindent \textbf{Modal Proposition Classification (MPC)}
In logic, modal propositions are divided into alethic modality, cognitive modality, deontic modality and dynamic modality \citep{wu2021semantic}. 
We merged the alethic modality and cognitive modality into  possible modality as mentioned before.
This task is to predict the category of a given modal proposition. We define it as a three-category task (\textit{possible/deontic/dynamic}).

\section{Dataset Construction}

\subsection{Data Acquisition}

\noindent \textbf{None-modal proposition}
ProPC \citep{hu2021propc} is a dataset constructed for none-modal proposition. We used it as the data source for relabeling and partitioning. For cross-domain data, we chose three common fields for research, namely medical, law and finance.
There is some overlap between the different fields where data is divided in ProPC, for example, data in the news field may include law and financial fields. To address this issue, the data in the news field has been reclassified into three other domains.
In addition, ProPC did not consider modal proposition during construction, resulting in some modal propositions being included. Sentences containing modal words are extracted and kept as corpus for modal proposition.

\noindent \textbf{Modal proposition}
Most statements in encyclopedia are deterministic statements describing facts. There are many modal events and relatively complete sentences in novels, blogs, and publications. We extracted this part of sentences from the CCL corpus \citep{CCL} as our original corpus.
We selected 12 typical modal words \citep{wu2021semantic} as trigger words with 1,000 statements of each as comprehensive corpus.
In the same way,  we found 100 statements corresponding to each keyword in three domains of finance, law, and medical.
\noindent \textbf{Categorical proposition}
Categorical proposition is a simple proposition. Statements in encyclopedia are mostly nested logical relations and positive semantics.
In order to make the corpus contain as many sentences as possible that are sufficient for all types,
we added the statements from LogiQA \citep{liu2020logiqa},
which is collected from publicly available questions of the National Civil Servants Examination of China.
To obtain sufficient categorical propositions and negative-semantic statements, we also 
removed the statements containing the logical keywords of compound propositions and focused on extracting some statements with negative words in the predicate part.
For the corpus of vertical domains, we use sentences from THUNews \citep{THUNews} and corresponding domain in the non modal propositional corpus for labeling.

\begin{table}[t]
\centering
\resizebox{\linewidth}{!}{
    \begin{tabular}{c c c c c c} 
    \hline
    Task & C\&E & FN & Law & Med \\
    \hline
    MPR & 23,948 & 2,006 & 1,957 & 2,067 \\
    NPC & \makecell[c]{6,659(EK)\\553(EI)} & 386   & 537   & 477   \\
    MPC & 13,935 & 1,230 & 1,231 & 1,350 \\
    CPC & 10,000 & 1,000 & 417 & 537 \\
    \hline
    \end{tabular}
}
\caption{Overview of the propositions contained in the dataset for each task.}
\label{table:tasks data}
\end{table}

\begin{table}[t]
    \centering
    \resizebox{\linewidth}{!}{
    \begin{tabular}{c c c c c c} 
        \hline
        Type & EK & EI & FN & Law & Med \\
        \hline
        Category & 3,794 & 120 & 42 & 234 & 183 \\
        Conjunctive & 1,569 & 367 & 258 & 218 & 262 \\
        Hypothetical & 1,082 & 56 & 85 & 81 & 28 \\
        Disjunctive & 213 & 10 & 1 & 4 & 4 \\
        Not & 2,542 & 258 & 389 & 188 & 239 \\
        Total & 9,201 & 812 & 776 & 726 & 717 \\
        \hline
    \end{tabular}
    }
    \caption{The overview of propositions in non-modal dataset.
    "EK" refers to encyclopedia data with logical keywords (not all explicit propositions). "EI" refers to encyclopedia data that conforms to natural language distribution (including implicit propositions). "Not" refers to sentences that are not propositions.
    }
    \label{table:non-modal overview}
\end{table}

\subsection{Data Annotation}

Given that some expressions in Chinese omit logical constants, the basic principle of annotation is to focus on semantic logic rather than keywords.
The annotation was conducted by 9 undergraduate and postgraduate students with natural language processing and linguistic related professional knowledge background.
There are 3 stages of annotation: annotation training, trial annotation and formal annotation.
Annotators first received annotation training, which provided project background introduction and annotation specification explanation.
In the trial annotation, 50 sentences per category were used to test the annotator's performence. We provided feedback to help them calibrate the annotation criteria. For the annotators whose accuracy rate is less than 80\%, re-train and re-prepare the data for a new round of trial annotation quality inspection.
The consistency test results indicated that there was good consistency among the annotators (Fleiss Kappa=0.729, 0.652, 0.793) \citep{Fleiss1971Measuring}, and formal annotation can be carried out.
In the formal annotation, each sentence was annotated by two different annotators. If their results were inconsistent, a third annotator would be asked to re-annotate and discuss the case with the first two annotators to reach a consensus.

Finally, we obtained a total of 45k annotated sentences including 12k non-modal propositions, 20k modal propositions, and 12k categorical propositions.
Detailed data distribution is shown in Tables~\ref{table:tasks data}.
In this paper, we use "C\&E" to denote Comprehensive and Encyclopedia data, "FN" to denote Finance data, "Law" to denote Law data, and "Med" to denote Medical data.

\begin{table}[t]
    \centering
    \resizebox{\linewidth}{!}{
    \begin{tabular}{c c c c c} 
        \hline
        Type & C\&E & FN & Law & Med \\
        \hline
        Possible & 5,006 & 289 & 269 & 639 \\
        Deontic & 5,018 & 625 & 827 & 451 \\
        Dynamic & 3,911 & 316 & 135 & 260 \\
        Not & 2,640 & 134 & 162 & 100 \\
        Total & 16,575 & 1,364 & 1,393 & 1,450 \\
        \hline
    \end{tabular}
    }
    \caption{The overview of propositions in modal dataset.}
    \label{table:modal overview}
\end{table}

\begin{table}[t]
    \centering
    \setlength{\tabcolsep}{3.3mm}{
    \begin{tabular}{c c c c c} 
        \hline
        Type & C\&E & FN & Law & Med \\
        \hline
        A & 7,562 & 782 & 733 & 751\\
        E & 1,341 & 125 & 162 & 56\\
        I & 861 & 89 & 86 & 177\\
        O & 236 & 4 & 19 & 16 \\
        Total & 10,000 & 1,000 & 1,000 & 1,000\\
        \hline
    \end{tabular} 
    }
    \caption{The overview of propositions in categorical dataset. 
    }
    \label{table:categorical overview}
\end{table}

\begin{figure}[t]
    \centering
    \includegraphics[width = \linewidth]{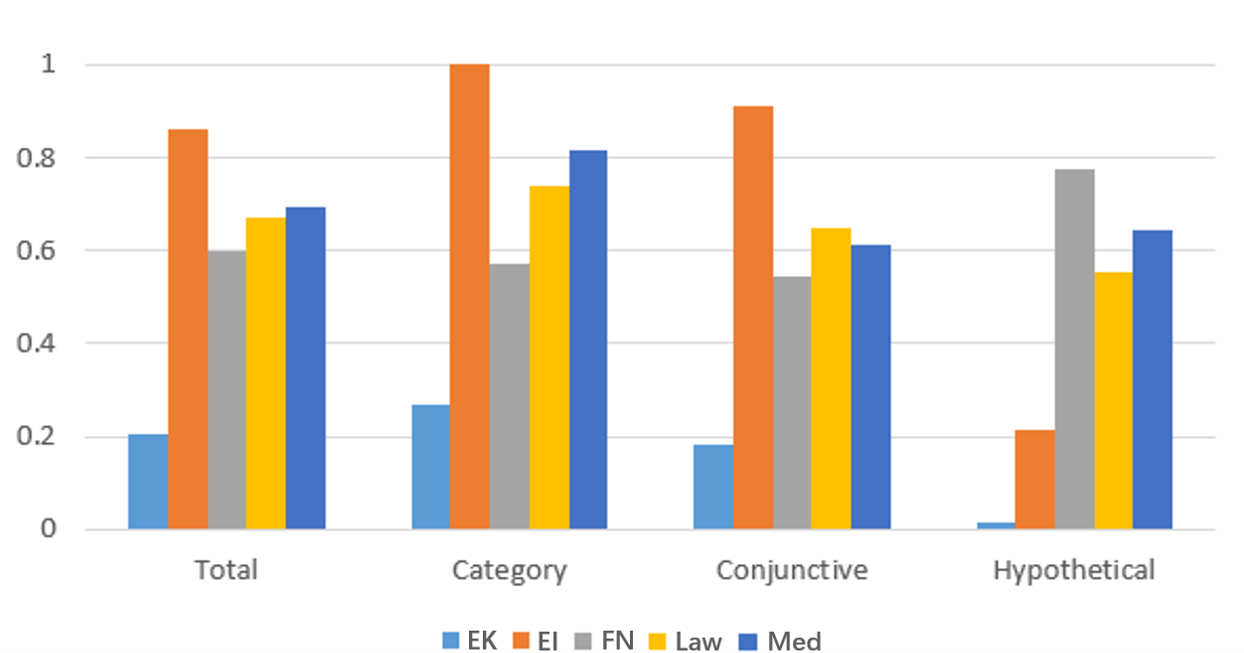}
    \caption{The distribution of implicit propositions in non-modal. Disjunctive proposition has no implicit form \citep{huang1991}.}
    \label{fig:implicit non-modal}
\end{figure}

\begin{figure}[t]
    \centering
    \includegraphics[width = \linewidth]{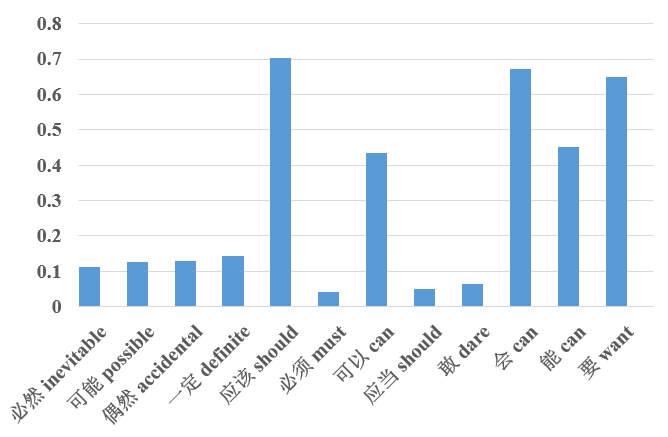}
    \caption{Cross-modal Keyword Proportion in C\&E}
    \label{fig:proportion_crossmodal}
\end{figure}

\subsection{Dataset Analysis}

\noindent \textbf{None-modal proposition}
The overall distribution of non-modal proposition dataset is shown in Table \ref{table:non-modal overview}. Non-modal proposition exists quite extensively in natural language.
As shown in Figure \ref{fig:implicit non-modal}, implicit non-modal propositions account for a considerable proportion in each domain.\footnote{Implicit propositions account for nearly 100\% of the categorical propositions in "EI". This is because "EI" data comes from naturally distributed encyclopedia data, and people's daily expressions are often implicit propositional forms.} While comparing the categories of implicit proposition, it can be found that implicit hypothetical propositions occur less frequently than the other two types under the natural distribution.
\noindent \textbf{Modal proposition}
Except for explicit propositions containing keywords, we also selected implicit modal propositions from sentences that don't contain keywords in the ProPC dataset for each domain.
From Table~\ref{table:modal overview}, we observed that sentences containing keywords account for a large proportion of modal propositions. "Not" means the sentence isn't a modal proposition, but it doesn't mean that these sentences are not propositions that need a further annotation.
During annotation, we noticed the phenomenon of cross-modal keywords. 
Specific cross-modal keywords distribution were given in Figure~\ref{fig:proportion_crossmodal}. Each keyword has cross-modal meanings. Some keywords such as "\begin{CJK}{UTF8}{gbsn}应该\end{CJK} (should)", "\begin{CJK}{UTF8}{gbsn}可以\end{CJK} (can)", "\begin{CJK}{UTF8}{gbsn}会\end{CJK} (can)", "\begin{CJK}{UTF8}{gbsn}能\end{CJK} (can)" and "\begin{CJK}{UTF8}{gbsn}要\end{CJK} (want)" have obvious cross-modal meanings.

\noindent \textbf{Categorical proposition} 
As shown in Table \ref{table:categorical overview}, "A" occupies the largest proportion in categorical dataset, which indicated that Universal affirmative proposition is most likely to appear due to their simplicity and directness.

\section{Experiments}
\subsection{Baseline Methods}
In order to explore and analyze the performance of machines in PEACE, we chooses several text classification methods as follows: Rule-based method, Majority, SVM \citep{2017Deep},
BERT \citep{devlin-etal-2019-bert}, and RoBERTa \citep{2019RoBERTa}. Besides, ChatGPT, a question-and-answer dialogue system based on a large language model, has made remarkable progress in natural language understanding. We also tested it's performance on our dataset.

\subsection{Experiments Setup}
Proposition classification task is a multi-class classification problem. Given a sentence-level text \begin{math}\mathcal{T}\end{math}and a set of labels \begin{math}\mathcal{L}=\left(l_{1}, l_{2}, \ldots, l_{n}\right)\end{math} for proposition classification, machine will learn a mapping \begin{math}\mathcal{C}: \mathcal{T} \mapsto \mathcal{P}(\mathcal{L})\end{math}. 
We use F1-score as the main evaluation metric, weighted across all classes.

\noindent \textbf{In-Domain Evaluation:}
We chose C\&E data as the dataset for exploring each in-domain task, and split it into training, validation, and testing set at an 8:1:1 ratio. For NPC tasks, we further tested the performance of models trained on explicit propositions on naturally distributed data (including implicit propositions), that is, to test the mapping of such a classifier \begin{math}\mathcal{C}({EK, EI})\end{math}.

\noindent \textbf{Cross-Domain Evaluation:}
We use C\&E data as \begin{math}\mathcal{T}_{source}\end{math}, and data of Financial, Law and Medical domains as the \begin{math}\mathcal{T}_{target}\end{math}. 
In order to explore model's generalization ability, we use the model trained on \begin{math}\mathcal{T}_{source}\end{math} and test it directly on \begin{math}\mathcal{T}_{target}\end{math}, that is, train a classifier \begin{math}\mathcal{C}(source, target)\end{math}. To further probe model's migration ability, we first train the classifier on \begin{math}\mathcal{T}_{source}\end{math} and then finetune it on \begin{math}\mathcal{T}_{target}\end{math}, that is, train a classifier \begin{math}\mathcal{C}(finetune, target)\end{math}. We choose BERT, which performs well on in-domain setting, as the model of this part, and label it as BERT-trans in the experiments.

\begin{table}[t]
    \centering
    \setlength{\tabcolsep}{1.5mm}{
    \begin{tabular}{c c c c c c c } 
        \hline
        Task  & Model & \multicolumn{5}{c}{Domains}\\
        \hline
        \multirow{8}*{MPR} &    & \multicolumn{2}{c}{C\&E} & FN & Law & Med \\  
        ~ & Rule-based & \multicolumn{2}{c}{0.34} & 0.27 & 0.30 & 0.28 \\
        ~ & SVM  &\multicolumn{2}{c}{0.69} & 0.60 & 0.63 & 0.64 \\   
        ~& BERT  & \multicolumn{2}{c}{\textbf{0.86}} & 0.65 & 0.79 & 0.68\\
        ~& RoBERTa & \multicolumn{2}{c}{\textbf{0.86}} & 0.62 & 0.75 & 0.48\\
        ~& BERT-trans  &\multicolumn{2}{c}{-} & \textbf{0.76} & \textbf{0.81} & \textbf{0.71}\\
        ~& ChatGPT-0  &  \multicolumn{2}{c}{0.38} & 0.35 & 0.43 & 0.4\\
            ~& Majority  &\multicolumn{2}{c}{0.44} & 0.49 & 0.51 & 0.53\\
        \hline
        \multirow{8}*{NPC} &    & EK & EI &FN & Law & Med \\
        ~ & Rule-based & 0.60 & 0.32 & 0.14 & 0.14 & 0.22 \\
        ~ & SVM  & 0.49 & 0.24 & 0.50 & 0.38 & 0.56 \\ 
        ~& BERT  & \textbf{0.91} & 0.48 & 0.42 & 0.64 & 0.48\\
        ~& RoBERTa & 0.90 & \textbf{0.54} & 0.32 & 0.67 & 0.48\\
        ~& BERT-trans  & - & - & \textbf{0.77} & \textbf{0.74} & \textbf{0.78}\\
        ~& ChatGPT-0 & 0.51 &  0.18 & 0.23 & 0.24 & 0.31 \\
        ~& Majority  & 0.56 & 0.22 & 0.11 & 0.44 & 0.38\\
        \hline
        \multirow{8}*{MPC} &   & \multicolumn{2}{c}{C\&E} & FN & Law & Med \\  
        ~ & Rule-based & \multicolumn{2}{c}{0.48} & 0.55 & 0.46 & 0.50 \\
        ~ & SVM  & \multicolumn{2}{c}{0.71} & 0.44 & 0.66 & 0.57 \\ 
        ~& BERT  & \multicolumn{2}{c}{\textbf{0.83}} & 0.78 & \textbf{0.83} & 0.78 \\
        ~& RoBERTa  & \multicolumn{2}{c}{\textbf{0.83}} & 0.79 & 0.82 & 0.80 \\
        ~& BERT-trans & \multicolumn{2}{c}{-} & \textbf{0.81} & \textbf{0.83} & \textbf{0.83}\\
        ~& ChatGPT-0 & \multicolumn{2}{c}{0.40} & 0.28 & 0.44 & 0.47 \\
        ~& Majority & \multicolumn{2}{c}{0.36} & 0.51 & 0.67 & 0.47 \\
        \hline
        \multirow{8}*{CPC} &   & \multicolumn{2}{c}{C\&E}  & FN & Law & Med\\
        ~ & Rule-based & \multicolumn{2}{c}{0.58} & 0.41 & 0.31 & 0.38 \\
        ~ & SVM & \multicolumn{2}{c}{0.77} & 0.80 & 0.74 & 0.68 \\
        ~& BERT  & \multicolumn{2}{c}{\textbf{0.95}} & \textbf{0.97} & 0.89 & 0.85\\
        ~& RoBERTa  & \multicolumn{2}{c}{\textbf{0.95}} & \textbf{0.97} & 0.87 & 0.85\\
        ~& BERT-trans &\multicolumn{2}{c}{-} &  0.90  & \textbf{0.97} & \textbf{0.93} \\
        ~& ChatGPT-0  & \multicolumn{2}{c}{0.34} & 0.22 & 0.41 & 0.48\\
        ~& Majority  & \multicolumn{2}{c}{0.74} & 0.76 & 0.62 & 0.64\\
        \hline
    \end{tabular}
    }
    \caption{The weighted average F1-score under cross-domain setting.}
    \label{table:cross-domain overview}
\end{table}

\begin{figure*}[t]
    \setlength{\belowcaptionskip}{-0.5cm}
    \begin{minipage}[c]{0.5\linewidth}
        \includegraphics[width = \linewidth]{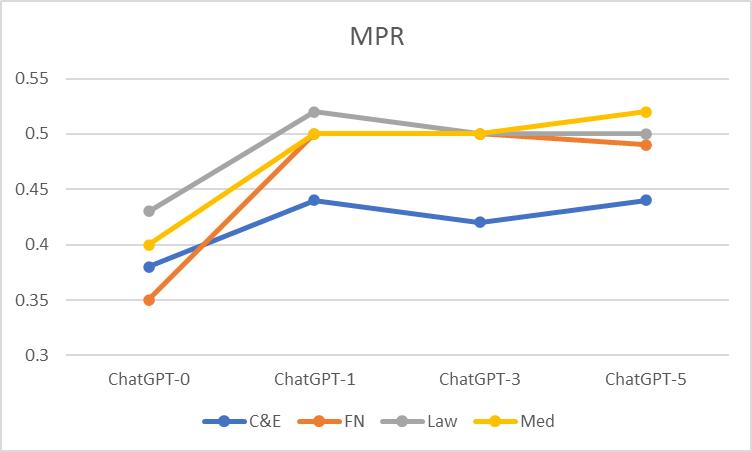}
    \end{minipage}
    \hfill
    \begin{minipage}[c]{0.5\linewidth}
        \includegraphics[width = \linewidth]{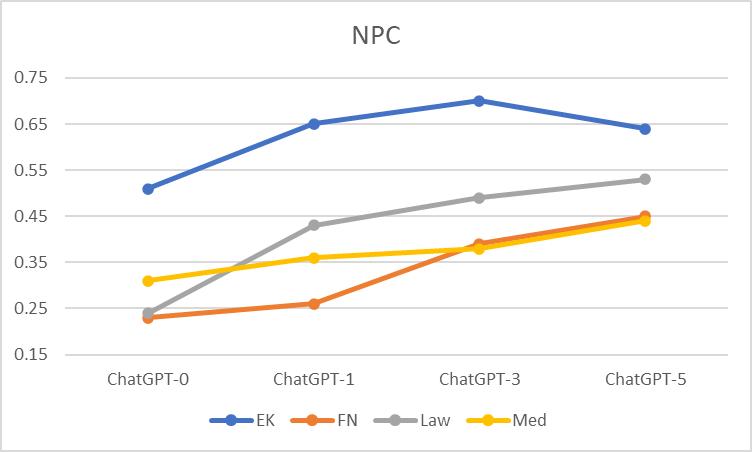}
    \end{minipage}
    \hfill
    \begin{minipage}[c]{0.5\linewidth}
        \includegraphics[width = \linewidth]{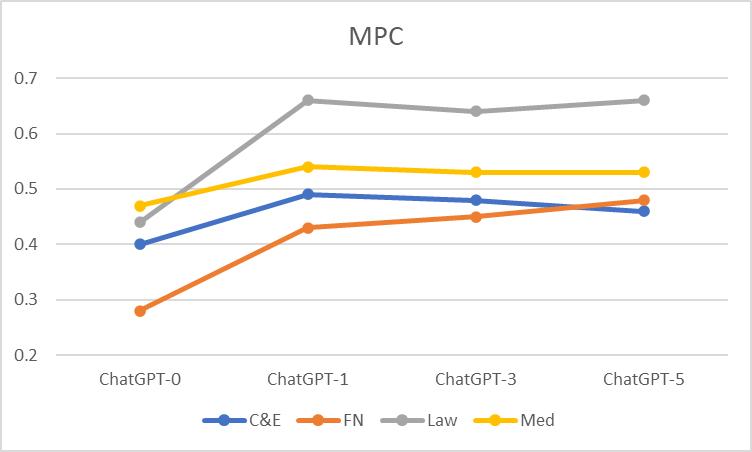}
    \end{minipage}
    \hfill
    \begin{minipage}[c]{0.5\linewidth}
        \includegraphics[width = \linewidth]{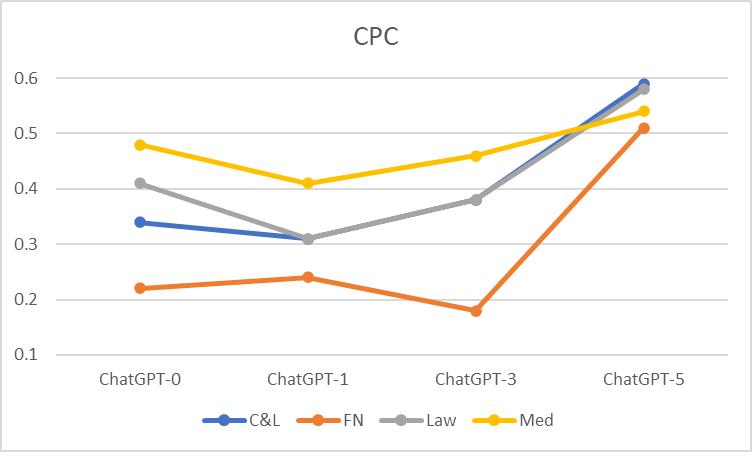}
    \end{minipage}
    \caption{The weighted average F1-score based on ChatGPT.
ChatGPT-0/1/3/5 represents providing corresponding cases for each category, where 0 represents direct testing.}
    \label{fig:ChatGPT}
\end{figure*}

\subsection{Results}
We evaluate model's performance on four tasks by in-domain and cross-domain settings. Results are shown in Table~\ref{table:cross-domain overview}.
\\
\\
\noindent \textbf{General Trends:}
Among all tasks, CPC task performed best overall in both in-domain and cross-domain settings. This may be due to more obvious characteristics of categorical proposition and the imbalanced distribution of data.
In each group of experiments, the results of Rule-based method were not ideal. This suggests it's not feasible to identify proposition only by logical keywords. Simultaneously constructing such a dataset conforming to natural distribution that contains implicit proposition has certain practical significance.
\\
\\
\noindent \textbf{In-Domain Evaluation:}
Pre-trained model BERT and RoBERTa achieved quite good results. After training on a large amount of corresponding data, the model is able to obtain propositional features and perform well in classification.
From the results of EK and EI in the NPC task, it can be seen that the model can learn the features of proposition classification from explicit proposition and partially transfer them to implicit proposition, but the effect is not good.
The MPC task performs slightly worse than the other three tasks, which may be related to the fact that a word has multiple modal meanings in the MPC task.

ChatGPT exhibits inferior performance compared to pre-trained models. By interacting with ChatGPT, we found that its definition of propositions still remains within the explicit concept strictly defined by formal logic and excels at capturing explicit rather than implicit propositional features, which cannot be well adapted to a wide range of practical situations.
We believe that this is because the task we propose is a brand new one, and ChatGPT has never seen such data or received similar training before. So the effect is not as good as BERT and RoBERTa who have been specially trained on the dataset. Even so, ChatGPT still outperforms rule-based methods on some tasks, such as the MPR task.
\\
\\
\noindent \textbf{Cross-Domain Evaluation:}
PLMs can be partially transferred to new domains for proposition classification tasks, with better performance than the majority of baseline models in \begin{math}\mathcal{C}(source, target)\end{math} setting. 
In the NPC task, rule-based method under cross-domain performs worse than in-domain due to more implicit proposition in cross-domain data, reflecting the challenge of applying explicit proposition classifier to it.
In the CPC task, the model performs better on finance due to differences in datasets.
The financial data comes from news with more regular sentence structure, while the data in medical and law fields comes from question-and-answer corpus. And colloquial datasets are more challenging. 
Comparing two settings of \begin{math}\mathcal{C}(source, target)\end{math} and \begin{math}\mathcal{C}(finetune, target)\end{math}, \begin{math}\mathcal{C}(finetune, target)\end{math} performs on par with or better than \begin{math}\mathcal{C}(source, target)\end{math}.
Among all tasks, results of NPC tasks are generally lower, implying that features of non-modal proposition are more difficult to learn.
ChatGPT performs relatively stable in both in-domain and cross-domain settings.

\subsection{Error Analysis and Discussion}

\begin{figure}[t]
    \centering
    \includegraphics[width = \linewidth]{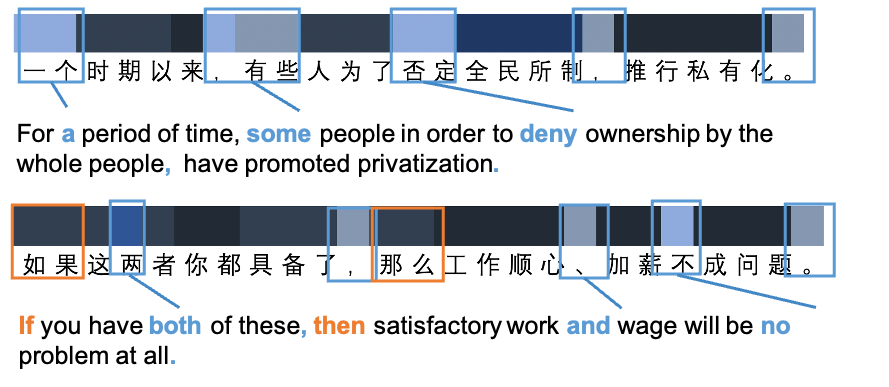}
    \caption{Examples of BERT attention visualizations in the categorical and non-modal proposition classification tasks, with categorical propositions at the top and non-modal propositions at the bottom.}
    \label{fig:error-non}
\end{figure}

\begin{figure}[t]
    \centering
    \includegraphics[width = \linewidth]{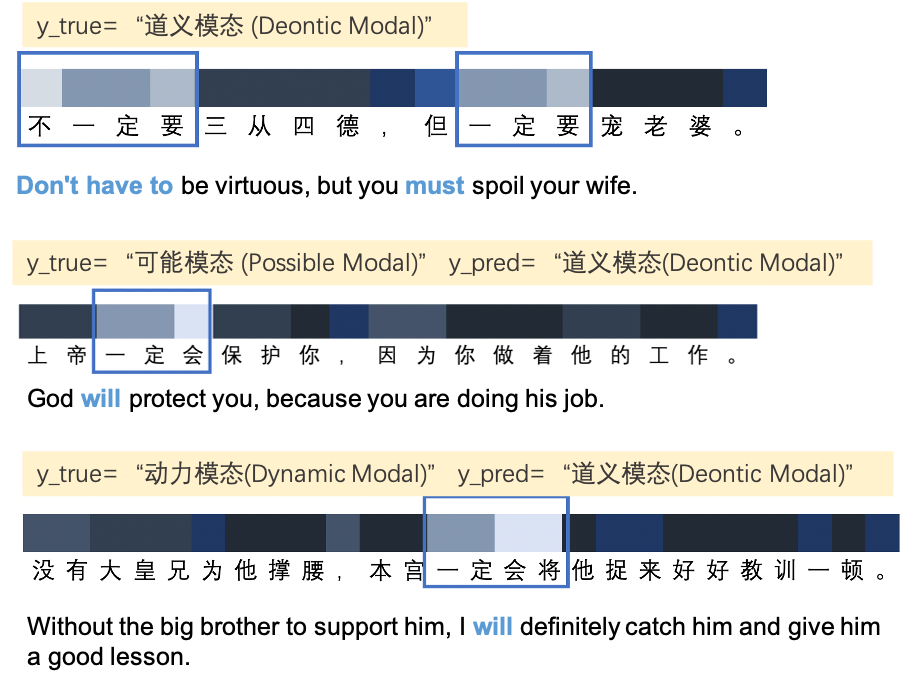}
    \caption{Examples of BERT attention visualizations in the modal proposition classification tasks.}
    \label{fig:error-modal}
\end{figure}

In order to deeply explore the proposition classification ability of ChatGPT, we provided different instructions (including definitions and examples) for it and conducted further experiments. See the Appendix ~\ref{sec:appendix} for specific instructions. The results are shown in Figure~\ref{fig:ChatGPT}. With the increasing number of cases provided to ChatGPT, it still shows an overall growth trend although its performance fluctuates slightly. This also demonstrates the effectiveness of our dataset.

For the pre-trained model, we performed attention visualization analysis based on BERT (see Figure~\ref{fig:error-non} and Figure~\ref{fig:error-modal} for details). For the part of the sentence that contributes the most to the result in a piece of data, the larger value of the corresponding position in the Attention matrix, the lighter color of the corresponding part in the figure. After analyzing some specific data,we found that the highlighted part in the figure overlaps with the logical keywords in the sentence, which indicates that the model effectively uses the knowledge information of some logical keywords when classifying. But there are also cases where some light colored parts have no connection with the parts related to the logical keywords in the sentence, such as punctuation and modal particles such as "\begin{CJK}{UTF8}{gbsn}的\end{CJK}".

\section{Conclusions}

In this paper, we introduce the concepts of explicit and implicit propositions based on the characteristics of Chinese, which is more suitable for Chinese NLP scenarios. To facilitate the research on it, we propose a comprehensive multi-level proposition classification system and create a large-scale Chinese proposition dataset PEACE from multiple domains, covering all categories related to propositions. Additionally, we conduct several evaluations on PEACE to evaluate the Chinese proposition classification ability of existing models and explore their limitations. Results show that BERT has relatively good proposition classification capability, but lacks cross-domain transferability. ChatGPT performs poorly, but its classification ability can be improved by providing more proposition information. Many issues are still far from being resolved and require further study.

\section*{Limitations}
Although domain migration is not difficult for humans, it is a major challenge for machines. For proposition classification tasks, although BERT has relatively good proposition classification ability, it doesn't perform well in cross-domain tasks. How to improve the domain transfer capability of models will be a good research direction in the future. In addition, there are currently many large language models, and we only selected a more representative LLMs ChatGPT for evaluation in this paper. In the future, we will explore the performance and limitations of more LLMs in this direction.

\bibliography{anthology,custom}
\bibliographystyle{acl_natbib}

\appendix

\section{Appendix}
\label{sec:appendix}
\subsection{Instruction for ChatGPT}

System\begin{CJK}{UTF8}{gbsn}：“你是一个有用的助手，记住并理解以下含义：” + 各类别命题定义 + “判断下面模态命题属于可能模态、道义模态、动力模态中的哪一类，不需要解释。”\end{CJK}
\textit{(You are a helpful assistant, remember and understand the following meanings:  + Definition of each category of propositions + Determine which category of possible, deontic, or dynamic modal propositions the following modal propositions belong to, without explanation.)}

Q: \begin{CJK}{UTF8}{gbsn}“这种暴发户意识，表现在消费行为上必然是畸形的高消费。”\end{CJK}
\textit{(This nouveau riche consciousness is inevitably manifested in abnormal high consumption behavior)}

A: \begin{CJK}{UTF8}{gbsn}“可能模态”\end{CJK}\textit{(possible modality)}

Q: ...		A: ...

...

\subsection{Co-occurence of Non-modal Proposition}
\begin{figure}[t]
    \centering
    \includegraphics[width = \linewidth]{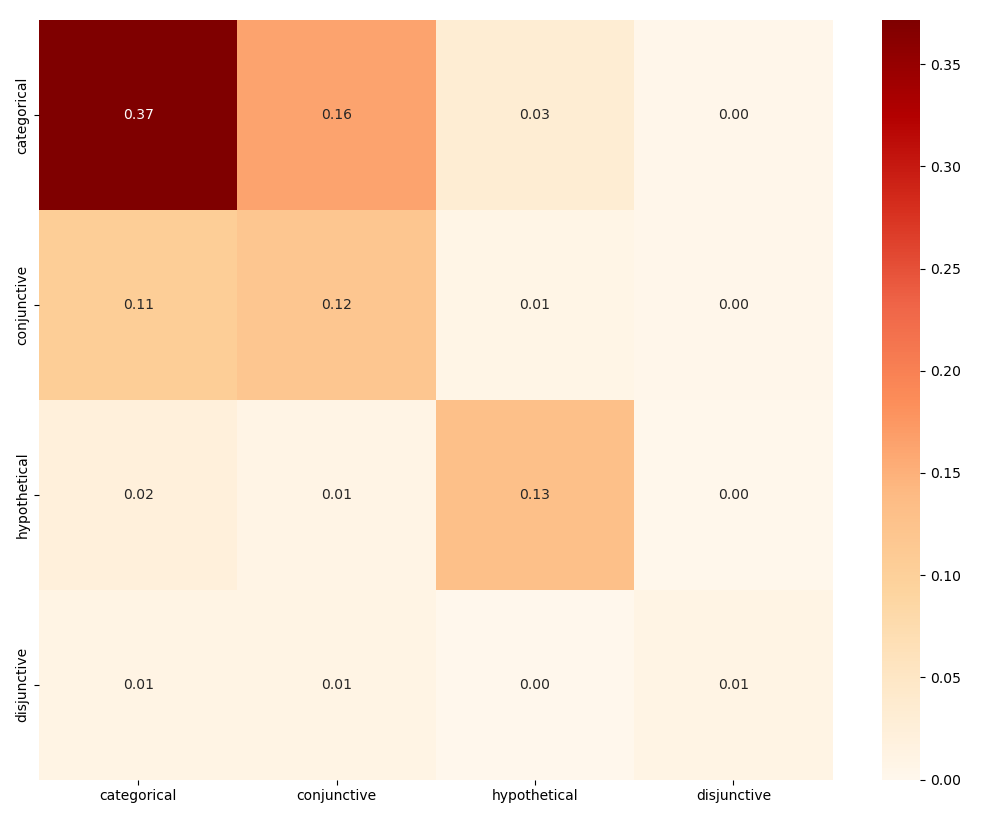}
    \caption{Caption}
    \label{fig:heatmap}
\end{figure}
Figure~\ref{fig:heatmap} shows the co-occurrence matrix of the classification on the C\&E dataset under the ChatGPT-5 setting. The value of 0.16 in the first row and second column of the figure indicates that 16\% of the data labeled as categorical propositions are classified as conjunctive propositions. From the figure, it can be seen that under the current setting, ChatGPT is prone to confusion between categorical proposition and conjunctive proposition.

\subsection{Examples of proposition}
We introduce the detailed examples for each category of proposition in Table~\ref{table:example of prosition}.
\begin{table*}[t]
\setlength{\belowcaptionskip}{-0.4cm}
\begin{center}
\begin{tabular}{c c} 
\hline
  Class & Example\\
  \hline
  \textbf{Categorical} & \\
   A & \makecell[l]{
   \begin{CJK}{UTF8}{gbsn}所有的鲸鱼都是哺乳动物。（显式）\end{CJK}  \textit{All} whales \textit{are} mammals. (explicit) \\
   \begin{CJK}{UTF8}{gbsn}正方形是矩形。（隐式）\end{CJK}  
Square is rectangular. (implicit) \\
   \hline
   }\\
    E & \makecell[l]{
   \begin{CJK}{UTF8}{gbsn}所有的行星都不是恒星。（显式）\end{CJK}
   \textit{All} the planets \textit{are not} stars. (explicit) \\
   \begin{CJK}{UTF8}{gbsn}人都不能知道自己去世之后的事情。（隐式）\end{CJK}\\
   People can't know what happens after they pass away. (implicit)  \\
   \hline
   }\\
    I & \makecell[l]{
   \begin{CJK}{UTF8}{gbsn}有的旁观者清。（显式）\end{CJK}  \textit{Some} bystanders \textit{are} aware of the situation. (explicit) \\
   \begin{CJK}{UTF8}{gbsn}当局者迷。（隐式）\end{CJK}  The player can not see most of the game. (implicit) \\
   \hline
   }\\
    O & \makecell[l]{
   \begin{CJK}{UTF8}{gbsn}有些人不珍惜生命。（显式）\end{CJK}  \textit{Some} people \textit{don't} value life. (explicit) \\
   \begin{CJK}{UTF8}{gbsn}人不珍惜生命。（隐式）\end{CJK}  People don't value life. (implicit) \\
   }\\ 
   \hline
  \textbf{Non-modal} & \\
   Categorical & \makecell[l]{
   \begin{CJK}{UTF8}{gbsn}\textit{所有的}人\textit{都是}贤良的。（显式）\end{CJK}  \textit{All} people \textit{are} virtuous.(explicit) \\
   \begin{CJK}{UTF8}{gbsn}人皆贤良。（隐式）\end{CJK}  Everyone is virtuous.(implicit) \\
   \hline
   }\\
    Conjunctive & \makecell[l]{
   \begin{CJK}{UTF8}{gbsn}小张不仅学问多而且很好学。（显式）\end{CJK}  \\Xiao Zhang is \textit{not only} knowledgeable \textit{but also} studious. (explicit)\\
   \begin{CJK}{UTF8}{gbsn}小张别说学问多了，压根就不好学。（隐式）\end{CJK}   \\Xiao Zhang, not to mention knowledgeable, doesn't like studying at all. (implicit)\\
   \hline
   }\\
    Hypothetical & \makecell[l]{
   \begin{CJK}{UTF8}{gbsn}如果要当一名合格的学生，那么就要好好学习。（显式）\end{CJK}  \\\textit{If} you want to be a qualified student, \textit{then} you have to study hard. (explicit)\\
   \begin{CJK}{UTF8}{gbsn}要当一名合格的学生就要好好学习。（隐式）\end{CJK}  \\To be a qualified student, one have to study hard. (implicit)\\
   \hline
   }\\
   Disjunctive & \makecell[l]{
   \begin{CJK}{UTF8}{gbsn}小王要么好好学习了，要么成不了好学生。（显式）\end{CJK}  \\Xiao Wang will \textit{either} study hard \textit{or} he won't be a good student.(explicit) \\
   }\\
   \hline
  \textbf{Modal} & \\
   Possible & \makecell[l]{
   \begin{CJK}{UTF8}{gbsn}水加温到了沸点\textit{必然}变成水蒸气。（显式）\end{CJK}\\  When water is heated to the boiling point, it \textit{must} become water vapor. (explicit) \\
   \begin{CJK}{UTF8}{gbsn}水加温到了沸点就得变成水蒸气。（隐式）\end{CJK}\\  When water is heated to the boiling point, it will change into water vapor. (implicit) \\
   \hline
   }\\
    Deontic & \makecell[l]{
   \begin{CJK}{UTF8}{gbsn}你\textit{应该}好好学习才行。（显式）\end{CJK}  You \textit{should} study hard. (explicit) \\
   \begin{CJK}{UTF8}{gbsn}你得好好学习才行。（隐式）\end{CJK}  You have to study hard. (implicit) \\
   \hline
   }\\
    Dynamic & \makecell[l]{
   \begin{CJK}{UTF8}{gbsn}我\textit{要}吃两碗饭。（显式）\end{CJK}  I \textit{want} two bowls of rice. (explicit) \\
   \begin{CJK}{UTF8}{gbsn}给我吃两碗饭。（隐式）\end{CJK}  Give me two bowls of rice. (implicit) \\
   }\\
\hline
\end{tabular} 
\caption{The examples of each class of proposition classification.} 
\label{table:example of prosition}
\end{center}
\end{table*}

\end{document}